\documentclass[a4paper, conference]{IEEEtran}
\usepackage{graphicx}
\usepackage{tabularx}
\usepackage{mathtools}

\usepackage{algorithm}
\usepackage{algorithmic}
\usepackage[numbers,sort&compress]{natbib}

\begin{document}

\title{Particle robots \\ A new specie of hybrid bio-inspired robotics}

\author{\IEEEauthorblockN{Luis A. Mateos }
\IEEEauthorblockA{http://www.particlerobots.com}

}

\maketitle
\thispagestyle{empty}
\pagestyle{empty}


\begin{abstract}

Inspired by a couple of simple organisms without eyes, neither ears. This paper presents a novel hybrid bionic robot, called "particle robot", which mix a macro-organism and a micro-organism in the same robot. On one hand, an interesting rather boring animal, the biological Echinoids (sea urchins) is mixed with the viruses micro-organisms, in specific the rotaviruses; together with spherical mobile robots. Analogously, from a pure robotic perspective, this bio-inspired robot can be seen as a spherical mobile robot wearing an actuated exoskeleton. 

The robot has two main configurations: when the spines are contracted it becomes a spherical mobile robot able to move in a fast pace on land, embedding all spherical mobile robots properties. 
On the other hand, when the spines or legs are extended in a controlled pattern, it can walk on flat surfaces as well as move on snow and over rocks as a bionic sea urchin. 

The spines of the robot are telescopic linear actuators, which combines soft and hard 3D print materials to make the actuation unit flexible for compressing it in minimal space and rigid for lifting the robot.


\end{abstract}

\section{Introduction}
\label{sec:introduction}


Hybrid bio-inspired robotics tries not only to mimic living organisms in nature. But also to improve them to make them adaptable to multiple terrains and environments, while mechanically modular and efficient. Further, bionic robots may embed the properties of more than one biological animal, plus the technological advances for improving the bio-inspired mixed version.

\subsection{Bio-inspired}

\begin{figure}[t]
	\centering
	\includegraphics[width=0.49\textwidth]{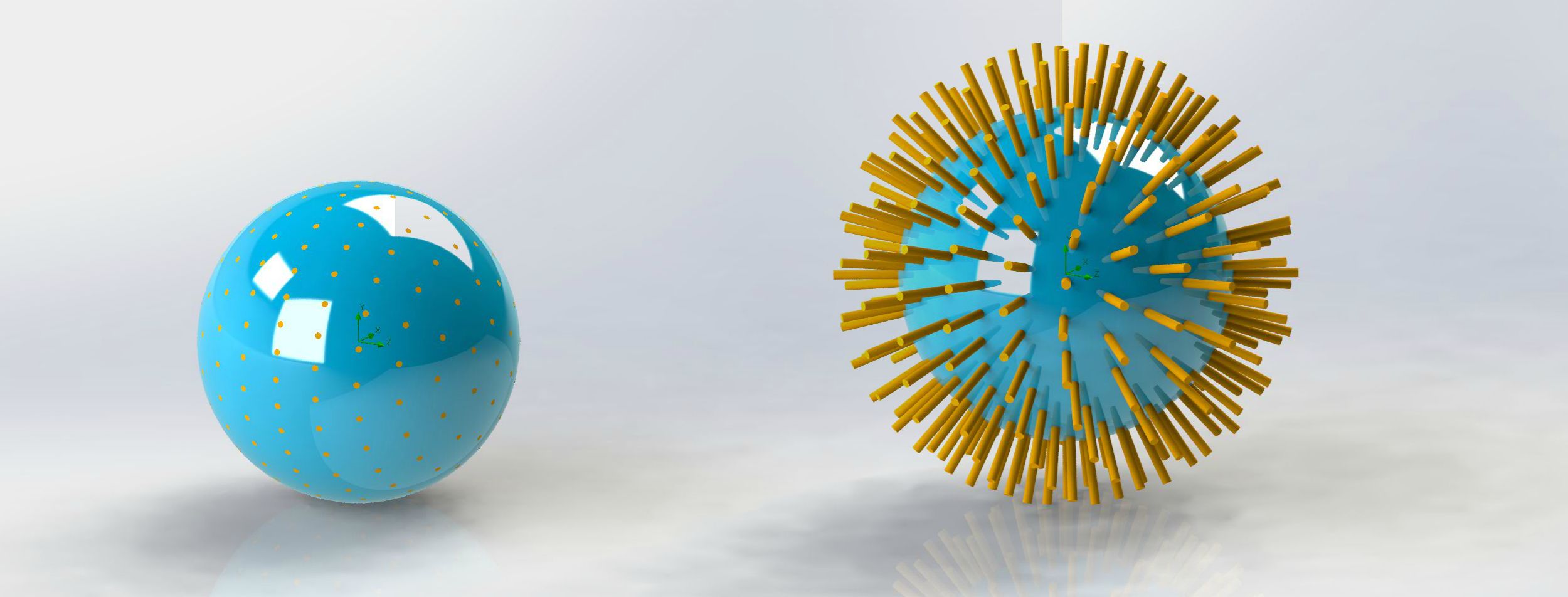}
	\caption{Particle robot. Spherical mobile robot configuration (left). Bionic robot (right). }
	\label{ffig1}
\end{figure}

The particle robot is inspired by two simple organisms which are very similar but exist at different scales: in the macro-organism-scale, the biological sea urchins and from the micro-organism-scale by the rotaviruses, together with spherical mobile robots, see Fig. \ref{ffig1}.

Sea urchins have a round shaped body and with long spines that come off it. The spines of the sea urchin are used for multiple purposes, such as protection, to move about, and to trap food particles that are floating around in the water. Sea urchins are animals with little mobility as they move slowly with their movable spines and typically range in size from $3$ to $10cm$ \cite{bookbarnes}.

\begin{figure*}[t]
	\centering
	\includegraphics[width=0.23\textwidth]{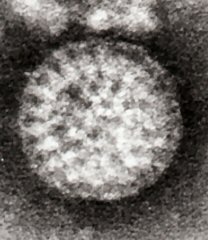}
	\includegraphics[width=0.38\textwidth]{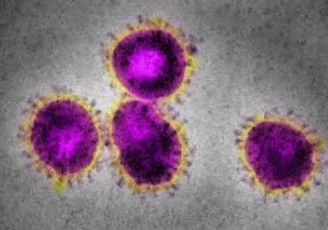}
	\includegraphics[width=0.2\textwidth]{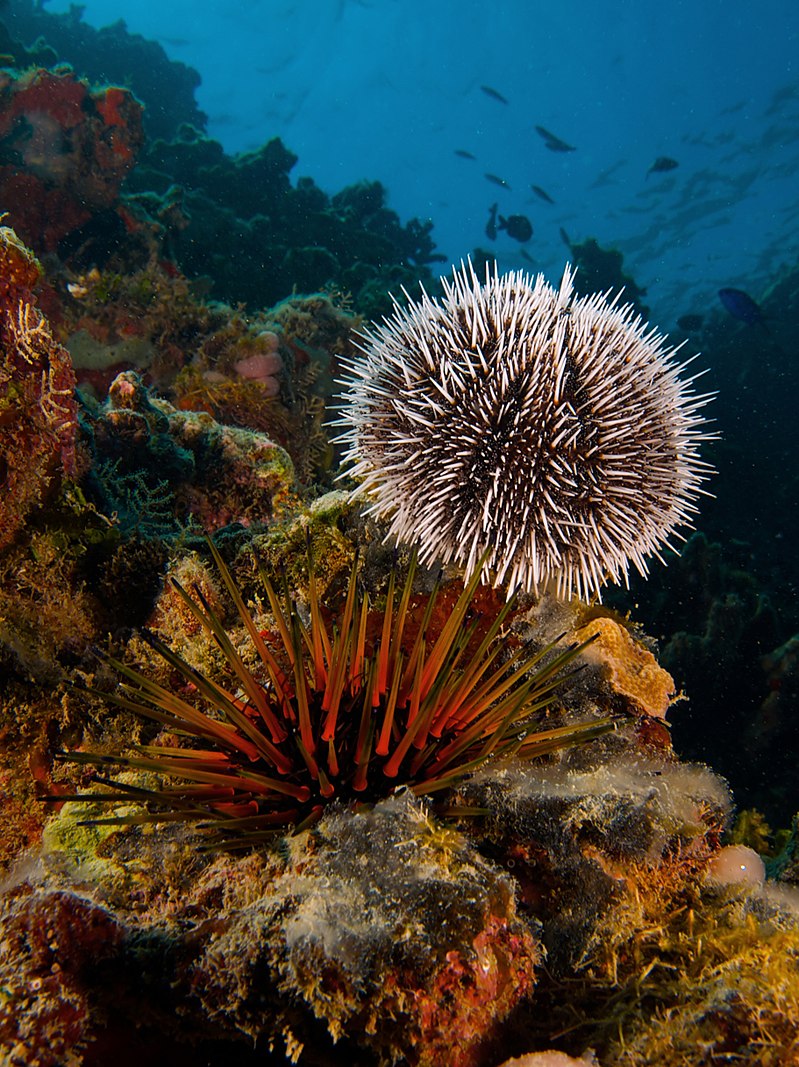}

	\caption{When examined by negative stained electron microscopy, rotaviruses often resemble wheels (left). Coronaviruses resembling spherical bodies with "legs" (middle). Tripneustes ventricosus sea urchin (top-right) and Echinometra viridis (bottom-right). (Images from wikipedia).}
	\label{ffig2}
\end{figure*}

Viruses exist in the realm of nanometers and can be found in different shapes such as "worms" like the Ebola virus, or more common shapes like wheels with "legs", i.e., HIV, influenza, corona-virus, among others, \cite{flewett}. The rotavirus type resemble a "wheel" ($rota$ in Latin) which can rotate and extend/compress the "legs" (Hemagglutinin and Neuraminidase) for connecting and propelling itself in the cell's 3D space, see Fig. \ref{ffig2}.


\subsection{Hybrid spherical robots}

Spherical mobile robots embed a special morphology that has multiple advantages over common legged and wheeled robots: their outer shell protects them and their motion is smooth with good power efficiency. In addition, these robots are omni-directional,  they can move in any direction as any part of their outer shell can be considered as a foot, making them easy to recover after a collision and automatic adapting to soft or uneven terrains \cite{Armour2006}.

Bio-inspired spherical mobile robots have been embedding mechatronic modifications to make them more adaptive to different terrains and environments, they can swim \cite{Suomela2005BallShapedRA}\cite{security}\, dive from integrated thrusters \cite{6618080}; move in snow from their rugged outer shell \cite{rotundus}; and even walk from using its shell as legs \cite{robotics1010003}; but never embedding an active exoskeleton. The main reason is the challenging mechatronic system to be fitted inside the constrained space of the robot's exoskeleton, which integrates a sealed spherical mobile robot as an inner shell together with a sensorized and actuated (exoskeleton) outer shell, plus managing their complex interactions. 

The exoskeleton is required to minimally increase the spherical robot's shell diameter, so the particle robot can still roll while embedding all the spherical robots properties. Otherwise, if the outer shell or exoskeleton diameter increases considerably, 
then the robot will not be able to roll when all legs are contracted. This can be seen as a pendulum-driven spherical mobile robot in which the pendulum is the inner-sealed spherical mobile robot and the outer shell is the particle robot itself. Hence, if the pendulum is very small it will not create the momentum for the robot to roll \cite{nizam}. 

The actuation system for the robotic spines requires a strong and fast extension able to lift the robot and to fit the compressed actuators inside the robot's spherical body.  
Therefore, a novel telescopic actuation is required for the particle robot, which requires minimal space when compressed and with a fast and strong extension to enable the robot to crawl with the spines.

\subsection{Electric telescopic actuators}

Electric linear actuators generally offer a low ratio between the length $L$ when fully extended to the length $l$ when compressed. Commonly, a fully extended actuator is less than twice the size when fully compressed $L<2l$. The reason is that in its body rest a rigid piston which can not be folded neither bended \cite{sclater2007mechanisms}. 

However, there has been other mechanical methods in linear actuators technology that can extend considerably while reducing their size when compressed. 
One example is the rigid chain actuator \cite{chainlink}. This is a specialized mechanical linear actuator used for push-pull lift applications. The actuator is a chain and pinion device that forms an articulated telescoping member to transmit traction and thrust. The links of the actuating member are linked in a way that they deflect from a straight line to one side only. As the pinions spin, the links of the chain are rotated $90^\circ$ through the housing, which guides and locks the chain into a rigid linear form effective at resisting tension and compression (buckling). In this way, the actuating member can be folded and stored compactly, either in an overlapping or coiled arrangement. 
However, one drawback of this mechanism is that it cannot support forces acting perpendicular to the axis of extension, since these will bend it in the same way as it is stored.

An improved version of the rigid chain actuator is the zip chain actuation unit, which interlocks a couple of articulated chains in a zipper-like fashion to form a single, strong column that enables push/pull operation with a maximum speed of 1000$mm/sec$. 
However, this actuation system requires a couple of articulated chains  \cite{zipchainlink}. 

Another highly compressible linear actuator is the Spiralift \cite{spiralift}, which is a system of interlocking horizontal and vertical metal bands that "unroll" to lift a load, this is the most compact electric linear actuator \cite{7487363}. The Spiralift can be integrated as a robot's vertical spine (eg. on EL-E \cite{Jain2009}, PR2\cite{pr2}, AMIGO\cite{6766556} and RED1\cite{qwerty}) -- increasing the robot's effective workspace and for easy transportation. 
Nevertheless, regardless of its highly compressibility, the Spiralift is slow to extend (max. speed of $10mm/s$) if compared to direct drive actuators, such as rigid chain actuator. The reason of the low speed is the interlocking process that rotates a band to form the tube. However, this mechanism is able to lock in all axes, and  withstand forces at any point of the extension.

\begin{table}[b]
	\centering
	\caption{Highly extendable linear actuators characteristics. }
	\begin{tabular}{|l|l|l|l|l|}
		\hline
		& \textbf{articulated }  & \textbf{speed}     & \textbf{locked}  \\ 
		& \textbf{ chain}  & \textbf{mm/s}    & \textbf{axes }  \\ \hline
		\textbf{Rigid chain}             & 1     & fast ($1000mm/s$)   & 2         \\ \hline
		\textbf{Rigid zip }              & 2     & fast  ($1000mm/s$)    & 3         \\ \hline
		\textbf{Spiralift}               & 1     & slow  ($10mm/s$)    & 3         \\ \hline
		\textbf{Articulated rack}       & 1     & medium  ($100mm/s$)   & 3         \\ 
		~      & ~    & (depends on motor and scale)  & ~         \\ \hline
	\end{tabular}
	\label{tabtable}
\end{table}

\begin{figure*}[t]
	\centering
	\includegraphics[width=0.99\textwidth]{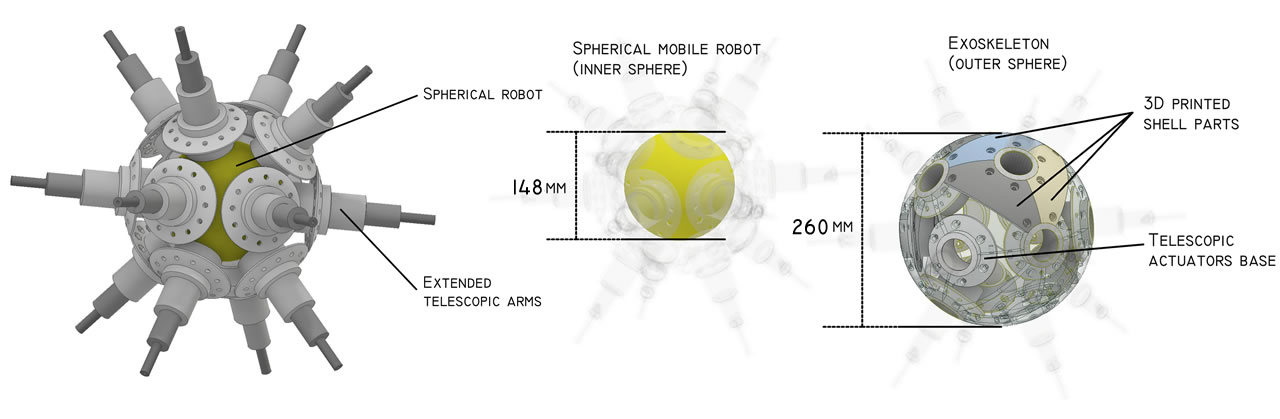}
	\caption{Particle robot 3D model with transparent outer shell (left). Spherical mobile robot as inner sphere (center). The exoskeleton or outer shell is made from 24 parts connecting to the telescopic bases (right).}
	\label{ffig3}
\end{figure*}

In this paper, a new kind of hybrid bio-inspired robotics is presented. The new specie called particle robots combines a macro-organism, the biological sea urchin animal and a micro-organism, the rotavirus; together with a spherical mobile robot. This new robotic breed embeds all the capabilities of spherical mobile robot properties and the dexterity to extend and compress its legs or "spines" as a bionic sea urchin robot for moving on different surfaces, such as snow and overcome obstacles. 

One of the challenges to develop this hybrid bionic robot is to integrate as many telescopic actuators as possible with the capability to compress them inside the spherical body of the robot, so it can still operate as a spherical mobile robot. Similarly, the telescopic actuators must be able to extend considerably, about half of its spherical body diameter, and be sufficiently strong to lift the robot up. 

This document is structured as follows: 
Section \ref{sec:design} presents the design of the particle robot and relevant features. 
Section \ref{sec:model} describes the simulations and locomotion modes of the robot. 
Conclusions are proposed in Section \ref{sec:conclusion}.


\section{Design}
\label{sec:design}


The development of hybrid robots requires new designs of mechanical elements that take advantages of 3D printer technologies combining soft and rigid materials in the same part for multi-functional purposes. The particle robot consists of three main parts: the inner shell or embedded spherical mobile robot, the outer shell or active exoskeleton and the 14 telescopic actuators as the legs or spines, see Fig. \ref{ffig3}. 

The particle robot can be remote control or semi-autonomous. The robotic system includes: xbee and blue-tooth modules, one STM32F103CBT6, a couple of micro-controllers ATmega328, seven DRV8835 dual motor driver, fourteen electromagnets for connectivity purposes and one BNO055 absolute orientation sensor. In this configuration, all 14 telescopic actuators can be controlled while registering the robot's inclination and orientation. In the semi-autonomous mode, the robot can be set for basic tasks: such as move in a straight line, stand in different legs and perform pre-programmed movements.

\subsection{Inner shell / spherical mobile robot}

Spherical mobile robots can be divided into three categories according to their driving mechanisms: direct-driving, gravity, and angular momentum.

In the direct-driving method the torque from the motor can be directly transmitted to the  shell as the locomotive force \cite{s1}\cite{s2}\cite{s3}\cite{s4}. The gravity method manipulates the position of the robot's center of mass to create a torque  with respect to the ground contact point to drive the robot to roll \cite{s5}\cite{s6}. In the angular momentum method a flywheel is installed inside the spherical robot. It rotates the robot's shell in the opposite direction to balance the angular momentum \cite{s7}.

The particle's spherical mobile robot integrates the direct-drive method, since the motor torques are directly transmitted to the robot's shells. Hence, the propulsion force is controllable if compared to the other methods and can be extended to a larger scale for fast locomotion and obstacle negotiation. In this configuration, the particle's spherical robot integrates two independent rubber-rimmed wheels inside the shell with a top slip bearing so the wheels are in firm contact with the inner shell and can communicate in blue-tooth and recharge wireless. 

\begin{figure*}[t]
	\centering
	\includegraphics[width=0.49\textwidth]{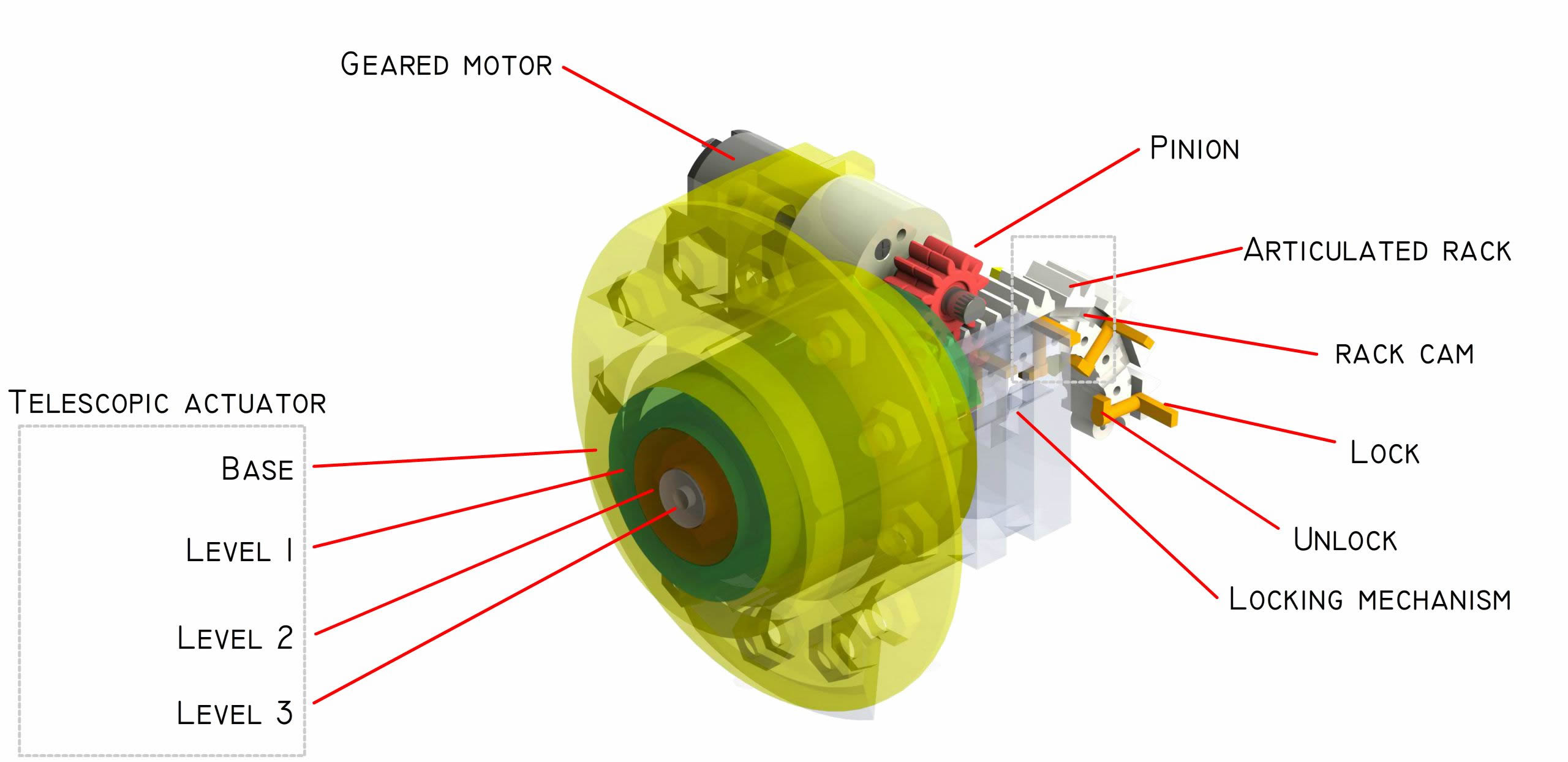}
	\includegraphics[width=0.41\textwidth]{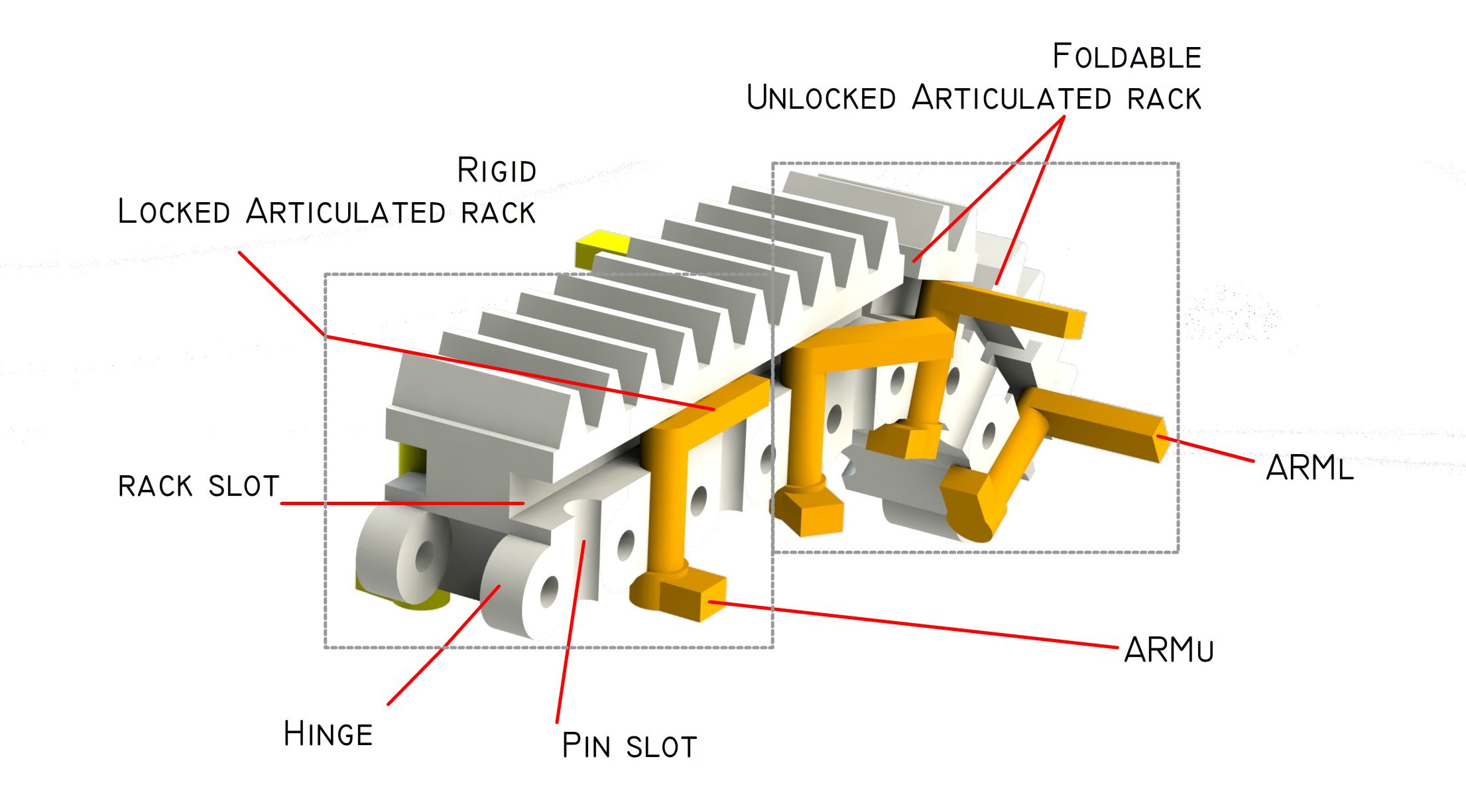}
	\includegraphics[width=0.4\textwidth]{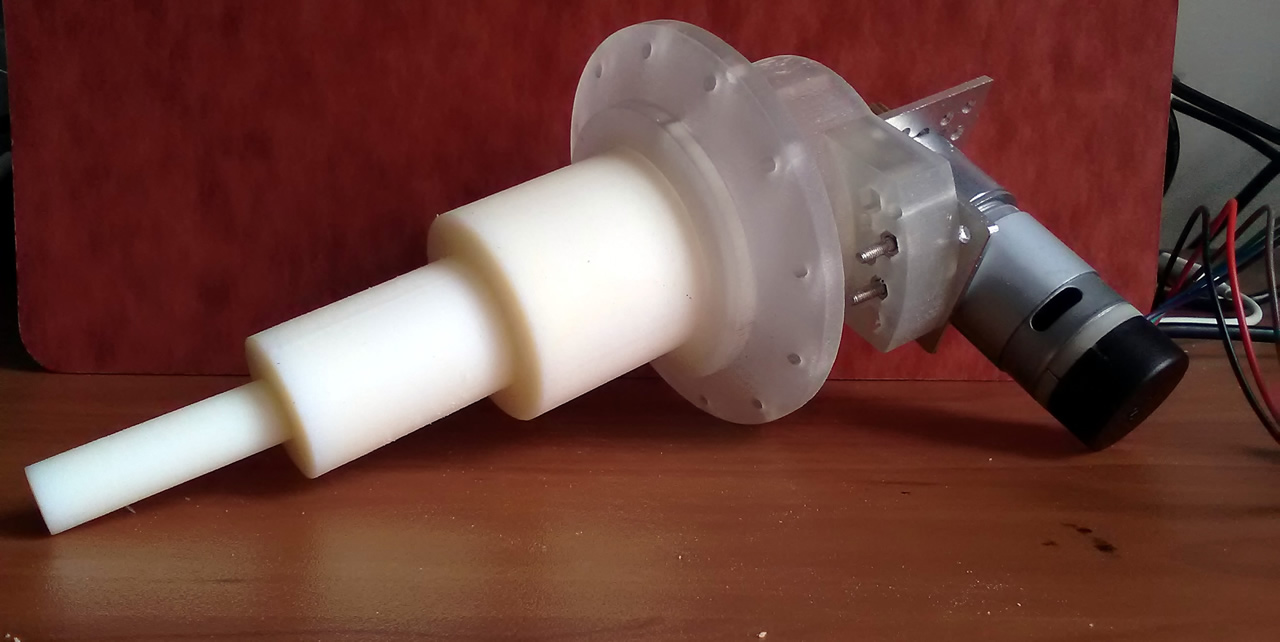}
	\includegraphics[width=0.4\textwidth]{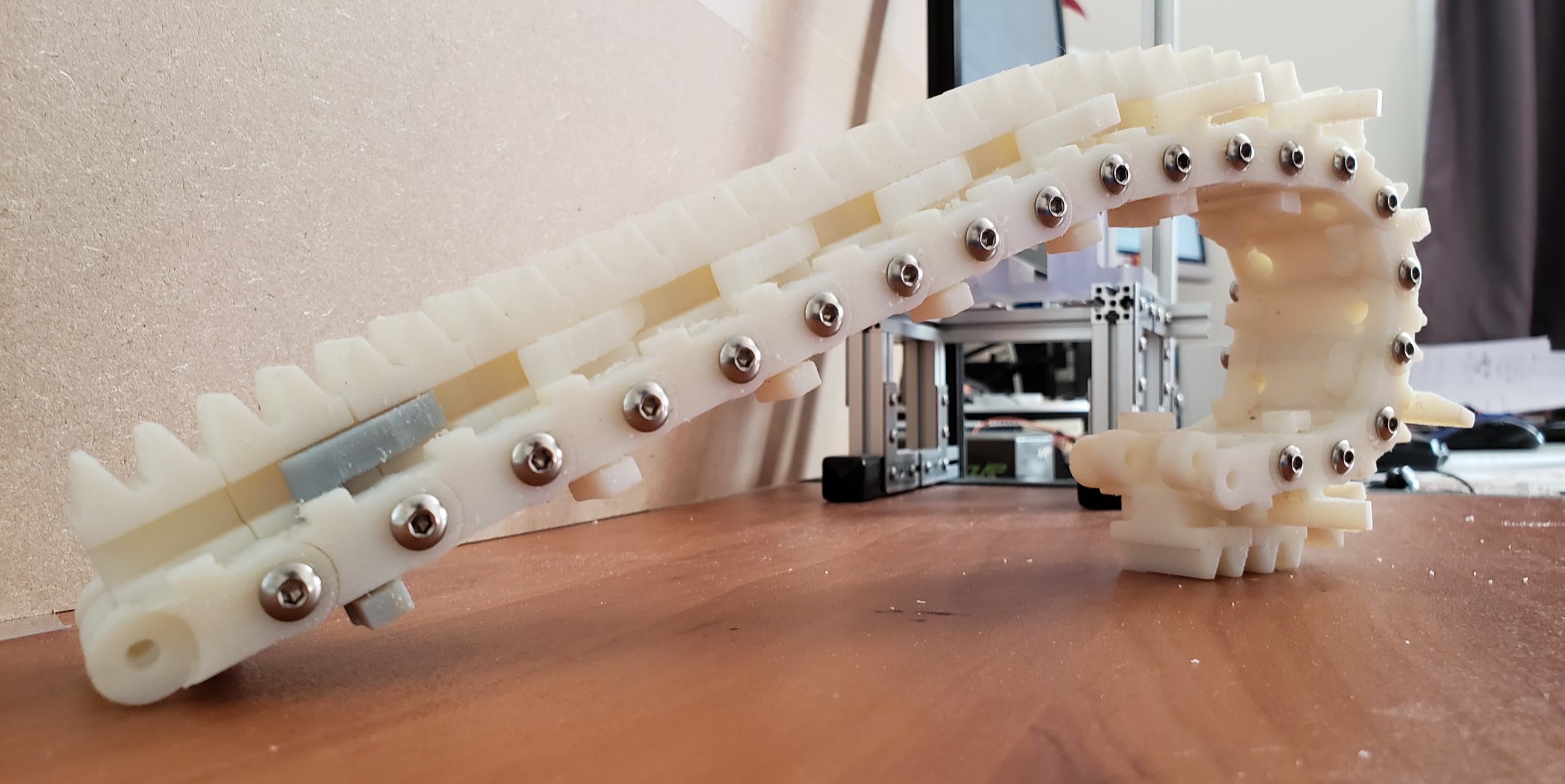}
	
	\caption{Telescopic actuator contracted in 3D model (top-left). Articulated rack components in 3D model. If the arm $ARM_L$ is set inside the rack slot, then the articulated rack becomes rigid. If the arm $ARM_L$ is rotated and set out of the rack slot, then the rack becomes articulated and can be folded. (top-right). 3D print telescopic actuator extended (bottom-left). 3D print articulated rack showing rigid and foldable modes in the same rack (bottom-left). }
	\label{ffig4}
\end{figure*}

\subsection{Exoskeleton / Outer shell}

The outer shell of the robot protects the electronics, motors and holds the telescopic actuators. It consists of 24 3D printed parts that forms a sphere with diameter $D_o=260mm$ and holds 14 telescopic actuators, see Fig. \ref{ffig3}. The material used for printing the shell is a mix of 3D print-materials (RGD450 as a rigid and TangoBlack as a flexible material) with a tensile strength of $30-35MPa$ and elongation at break 50-65\%. In this setup, the robot can withstand hits when rolling or falling from uneven surfaces. 

\subsection{Telescopic levels}

The telescopic actuator consists of four levels that can be extended up to $l=128mm$ from the top of its base. The length of the actuator when extended is $L=178mm$ and when contracted is equal to the $base_{height}=50mm$ with an extension ratio of 1:3.56. 

The base level of the telescopic actuator is 3D print with rigid material in order to support all the  telescopic levels and connections with the outer shell parts. 
The following three levels in the telescopic actuator are combinations of 3D printer materials: rigid $75\%$ and flexible $25\%$ in order to withstand hits, similar to the outer shell parts. 

\subsection{Articulated rack-gear}

The telescopic actuator integrates a 3D print rack \& pinion gearbox to extend and contract the levels. A geared motor attached to the telescopic' base level is the actuation unit "pinion", which drives the articulated rack attached to the top level of the telescopic arrangement. 

Each articulated rack consists of a couple of gear-teeth, a rack-slot on the sides, a pin with a couple of arms to lock/unlock the articulations and a hinge on both ends for linking the rack articulations, see Fig. \ref{ffig4}.

\subsubsection{Gear-teeth}

The articulations can be set with $n$ number of gear-teeth. However, when the articulations are folded, the more gear-teeth per articulation will lead to a lower degree of compression. 
The optimal solution for folding the articulated rack to a minimum dimension is with a single gear-tooth per articulation. Nevertheless, in the presented prototype, the articulations are set to a couple of gear-teeth. Since, it is easier to construct, arrange and still have a good compression ratio.

\subsubsection{Rack-slot and pin with arms}

Each articulated rack has a rectangular and cylindrical slots on each side. 
A pin is inserted in the cylindrical slot, this pin consists of a cylindrical segment with a couple of arms, each  located on each pin's edge. The arms are oriented $90^\circ$ from each other, one is always parallel to the rack extension axis and the other is perpendicular to it, pointing out from the rack gear. 
The arm $ARM_L$ is used for locking the articulated rack, while for unlocking the arm $ARM_U$ is used, see Fig. \ref{ffig4}. 

\subsubsection{Hinge on both ends}
The articulated racks integrate a hinge connector on both ends to interlock the articulations and create longer racks. The hinge connectors are also use for attaching a cover to the cylindrical pin with arms, so it is always trapped when rotating.

\subsubsection{Telescopic base}
The telescopic base supports all the telescopic levels, holds the gear-motor and integrates a guide for transforming the articulated rack from foldable to rigid and vice-versa. The holder integrates a couple of guides, one for each arm of the pin ($ARM_L$ and $ARM_U$) for locking or unlocking the rack articulations \cite{urchin}.

\subsubsection*{Articulated rack locking}  If the pinion attached to the gear-motor rotates clockwise, it will move the articulated racks out from the actuator, the telescopic-holder guides the pin's arm $ARM_L$ to fit inside the rack's slot, changing the articulation properties from foldable to rigid, while extending the telescopic actuator. 

\subsubsection*{Articulated rack unlocking} If the pinion rotates counterclockwise, it will move the articulated racks inside the actuator, so the guides from the telescopic holder rotates the $ARM_U$ and consequently rotating the $ARM_L$ releasing it from the rack's slot,  changing the articulation properties from rigid  to foldable \cite{videos}.

\section{Model and Simulation}
\label{sec:model}

%

\begin{figure}[t]
	\centering

	\includegraphics[width=0.24\textwidth]{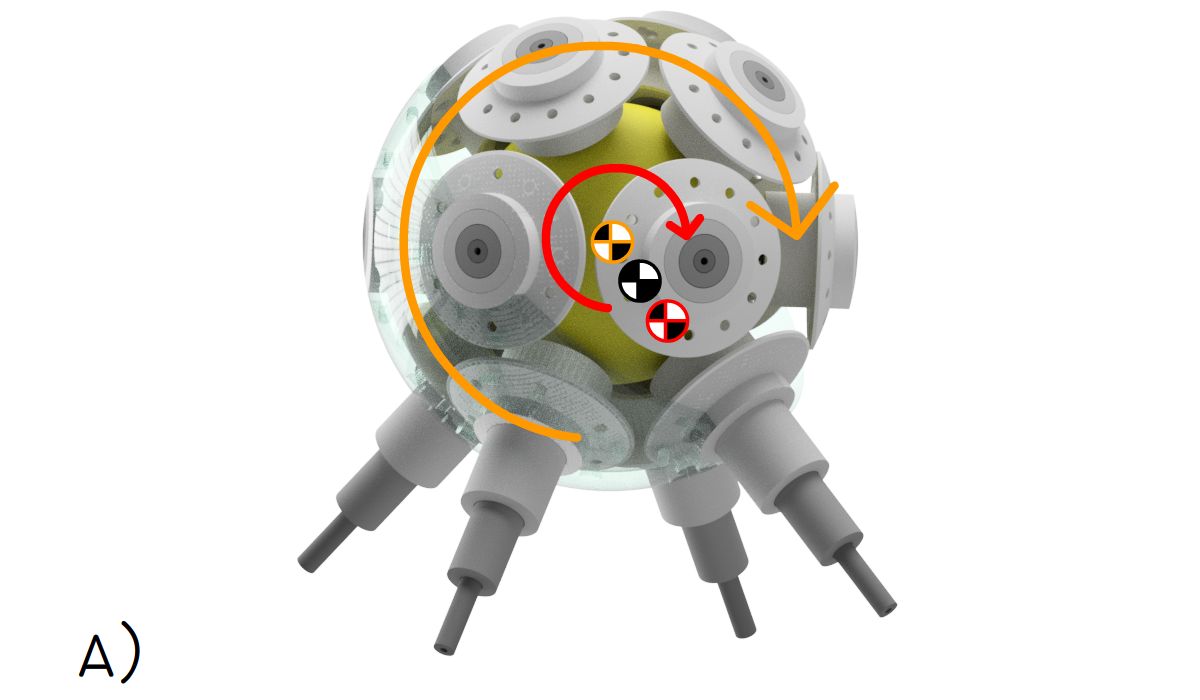}
	\includegraphics[width=0.24\textwidth]{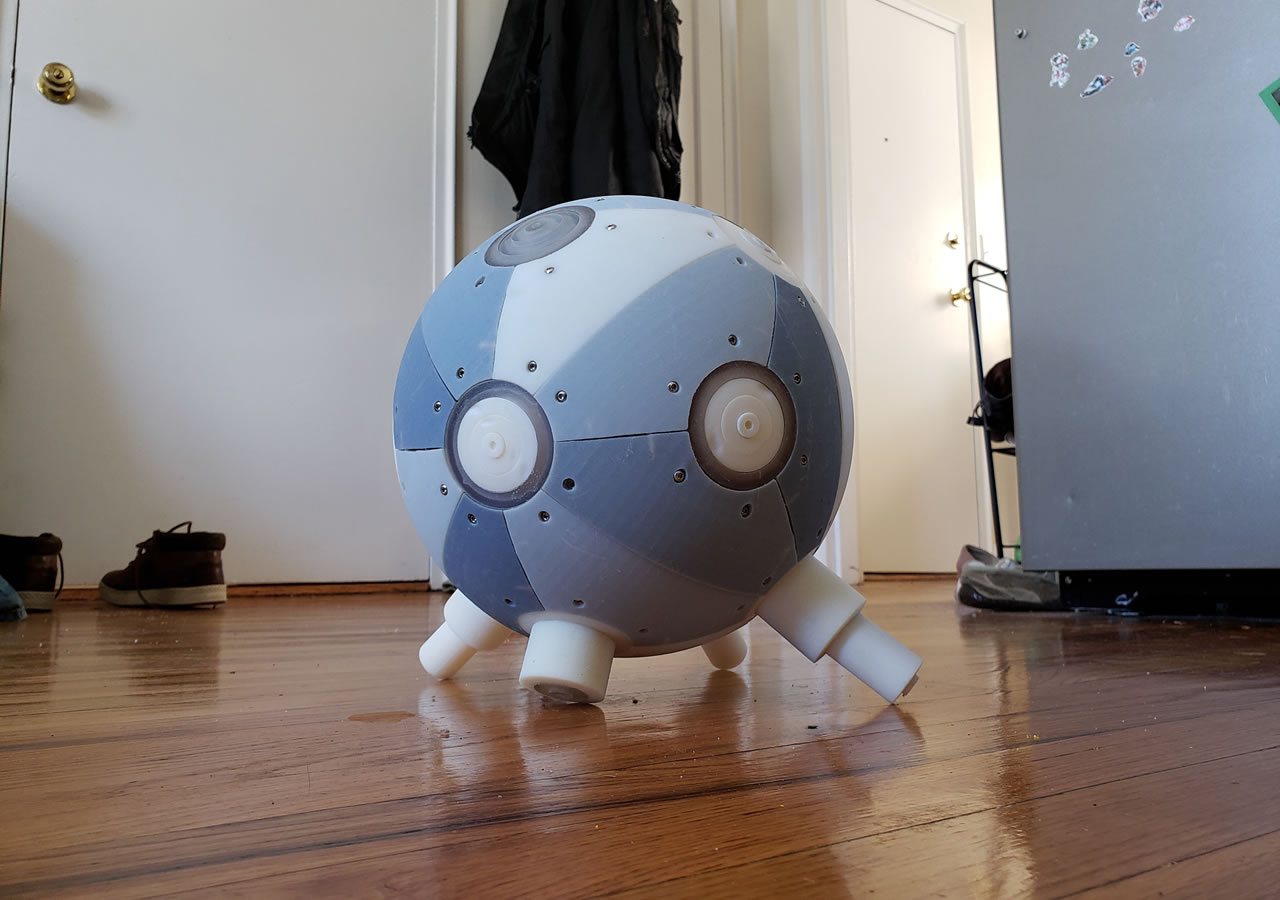}
	\includegraphics[width=0.49\textwidth]{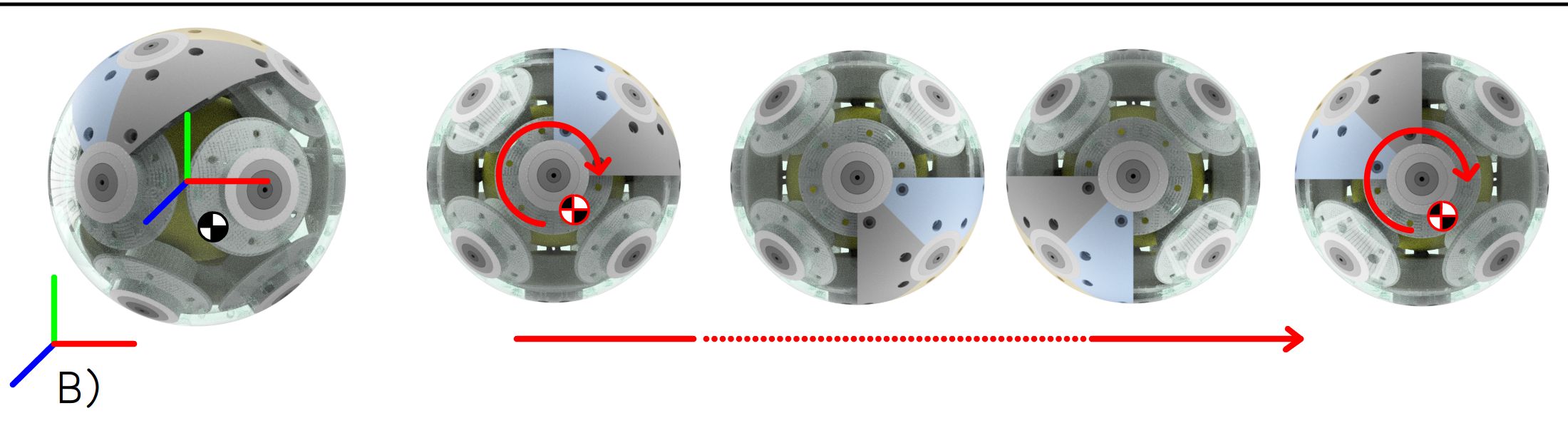}
	\includegraphics[width=0.49\textwidth]{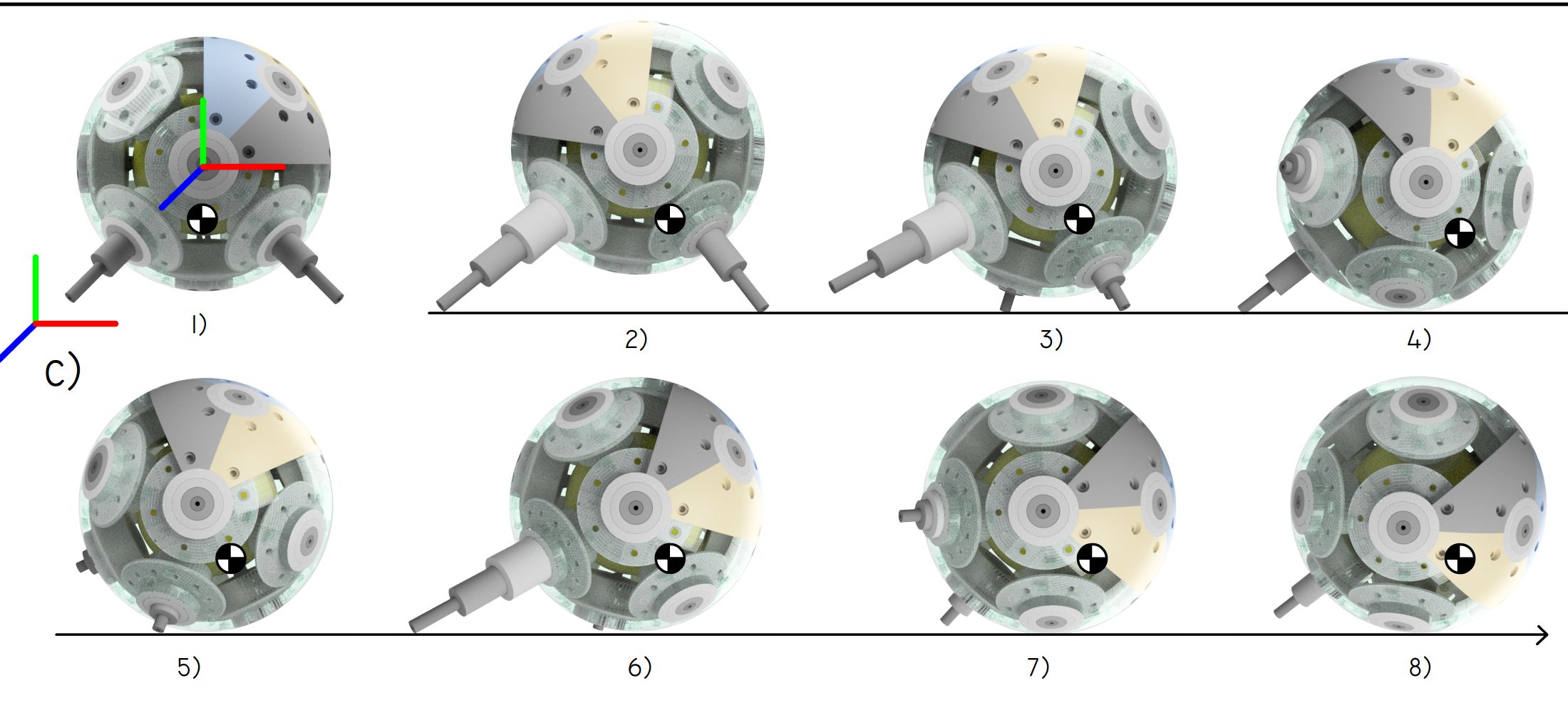}

	\caption{A) Particle robot 3D model and real prototype. B) The particle robot is able to roll from its inner spherical mobile robot. C) Extending and compressing the telescopic actuators to propel itself from a standing position. }
	\label{figlocomotion2}
\end{figure}

Common spherical mobile robots are suitable for rigid flat surfaces on which they can roll. In these environments, their shells are able to make proper contact to the surface and the locomotive forces from the inner robot are translated to the outer shell, making the robot roll.  
However, spherical mobile robots in challenging environment such as sand, snow or rocks, are unable to move. Since, their spherical body is not able to grip to these environments and as result the sphere is rotating but slipping out.

Spherical mobile robots are able to leverage these challenging environment by adding an outer rugged shell. 
This solution adds a passive element to the sphere without any actuation system, helping the robot to move on semi-rigid surfaces. 

In the evolution line of spherical mobile robots, the particle robot adds an actuated exoskeleton to the basic spherical mobile robot, making it more versatile and dexterous when facing extreme environments such as crud snow or rocks. Since, the robot combines rolling forces with walking patterns.


The particle robot has three locomotion modes: 1) Roll as a spherical mobile robot, 2) Walk by extending and compressing the spines in controlled pattern, 3) Combining walking and rolling.

\subsection{Rolling with inner spherical mobile robot}

The particle robot embeds the properties of spherical mobile robots when all the spines are compressed in its body. Therefore, the robot can rely only in the forces exerted by the inner spherical robot to start moving. 

Figure \ref{figlocomotion2}, shows a simulated particle robot and the basic locomotion mode of rolling.


\begin{figure}[b]
	\centering
	\includegraphics[width=0.49\textwidth]{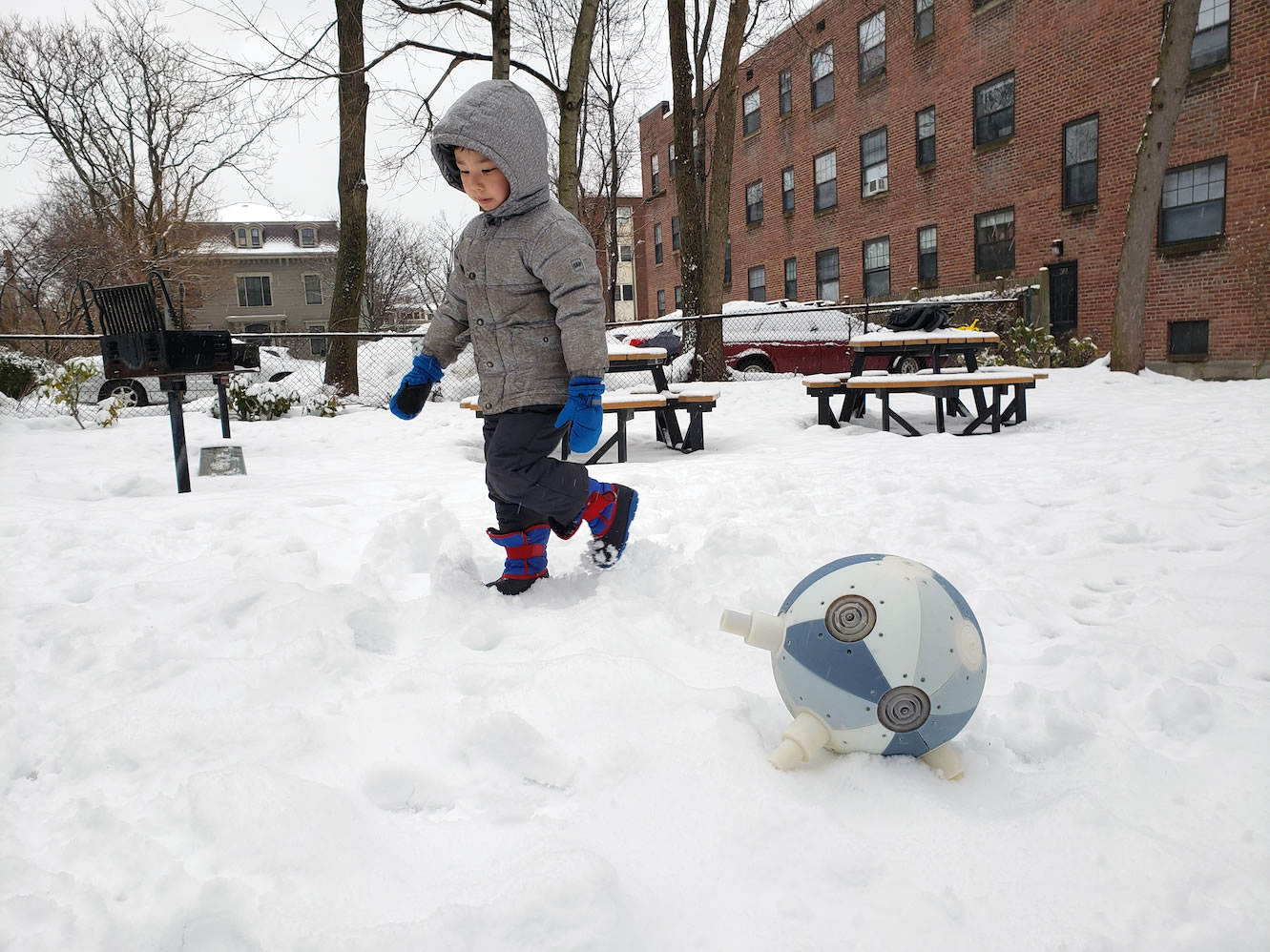}
	
	\caption{Particle robot in crud snow. }
	\label{lugos}
\end{figure}

\subsection{Walking with spines}

Similar to the bionic sea urchin robot \cite{urchin}, the particle robot is able to stand in 3, 4 or 5 spines and start moving by extending and compressing the spines in different patterns, see Fig \ref{figlocomotion2}.

\subsection{Walk and rolling}

In challenging environment, such as crud snow, where the snow gets packed in certain places, piled in others, generally an even uneven surface with slippery patches and huge lumps of powder. Spherical mobile robots fail to move. 

In these non-rigid environments, the spherical mobile robots with polished or rugged shells get jammed. This is due to the poor friction coefficient between the shell and the snow/ice. Similarly, small robots with legs will be unable to pass snow lumps, even though they are more versatile than spherical mobile robots.

On the other hand, the particle robot is able to move on rocky environments and different types of snow by combining the spinning inertia forces from its internal spherical mobile robot together with the walking behavior, see Fig. \ref{lugos}.

\begin{figure}[t]
	\centering
\includegraphics[width=0.45\textwidth]{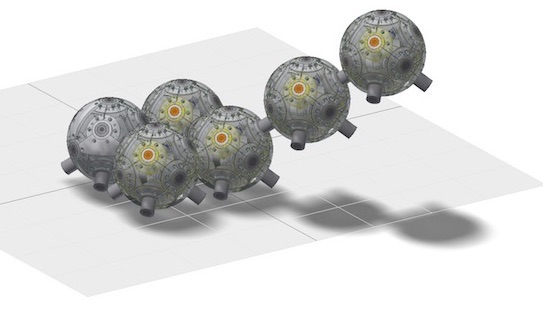}
	\caption{Particle robot prototypes (left). Virtual 3D robotic formation with particle robots (right). }
	\label{figfinal}
\end{figure}

\subsection{Swarm robotic formation in 3D}

Particle robots were simulated in an empty space environment with no gravity neither inertial forces. In this virtual conditions each particle robot can position and latch to other particles. The robots can move in 3D space and position in a 3D coordinate with a certain pose. The spines are able to latch to spines of other robots that are nearby. The particles were simulated moving in different frequencies and speeds, resulting in random pattern. 

In a realistic setup, the particle robot must be able to submerge and control its position by rotating with the inner spherical mobile robot, together with the spines for balancing purposes underwater. Further, the spines of the robot must integrate an electromagnet or additional mechanisms for latching and connecting to other particle robots. \cite{8793525}.









\section{Conclusion and future work}
\label{sec:conclusion}

%

In this paper, a new specie of hybrid bio-inspired robot is presented, called particle robot. Bio-inspired from micro and macro-organism, such as wheel-shaped viruses, rotaviruses and from the Echinoid (sea urchin) animal, but combined with a spherical mobile robot. This new robotic specie integrates in its body a spherical mobile robot as an inner sphere wearing an actuated exoskeleton or outer shell with the capability to extend and contract its spines (telescopic actuators). 
In this setup, the particle robot is able to perform all tasks from the spherical mobile robots when its spines are contracted; plus task as an Echinoid robot, such as move on snow, over obstacles and crawl on uneven terrain, when its spines are moved in controlled patterns.

Further, a novel telescopic actuation system is presented. It is able to extend considerably while contracting to a minimal space to fit inside the particle robot, in between the inner-spherical robot and the outer exoskeleton. The telescopic actuation combines rigid and flexible 3D printed materials to create a semi-rigid rack-pinion gearbox.

An important next step in the development of the particle robot is to make it waterproof IP68, protected from immersion in water with a depth of more than 1 meter. Since, the current prototype of the particle robot is not fully submersible, and the robots cannot be use for 3D robotic formations yet. One can image 10, 50 or more particle robots in the swimming pool moving at certain frequencies and coming together forming shapes we see in the nature, similar to swarm cells assembly or structures in the matter, such as ice crystals.

\addtolength{\textheight}{-3cm}   



\bibliographystyle{IEEEtran}
\bibliography{simple}

\begin{thebibliography}{10}
\providecommand{\url}[1]{#1}
\csname url@samestyle\endcsname
\providecommand{\newblock}{\relax}
\providecommand{\bibinfo}[2]{#2}
\providecommand{\BIBentrySTDinterwordspacing}{\spaceskip=0pt\relax}
\providecommand{\BIBentryALTinterwordstretchfactor}{4}
\providecommand{\BIBentryALTinterwordspacing}{\spaceskip=\fontdimen2\font plus
\BIBentryALTinterwordstretchfactor\fontdimen3\font minus
  \fontdimen4\font\relax}
\providecommand{\BIBforeignlanguage}[2]{{%
\expandafter\ifx\csname l@#1\endcsname\relax
\typeout{** WARNING: IEEEtran.bst: No hyphenation pattern has been}%
\typeout{** loaded for the language `#1'. Using the pattern for}%
\typeout{** the default language instead.}%
\else
\language=\csname l@#1\endcsname
\fi
#2}}
\providecommand{\BIBdecl}{\relax}
\BIBdecl

\bibitem{bookbarnes}
R.~D. Barnes, ``Invertebrate zoology.'' \emph{Philadelphia, PA: Holt-Saunders
  International.}, p. 961 – 981, 1982.

\bibitem{flewett}
W.~G. Flewett, T.H., ``The rotaviruses.'' \emph{Archives of Virology 57}, p. 1
  – 23.

\bibitem{Armour2006}
\BIBentryALTinterwordspacing
R.~H. Armour and J.~F.~V. Vincent, ``Rolling in nature and robotics: A
  review,'' \emph{Journal of Bionic Engineering}, vol.~3, no.~4, pp. 195--208,
  Dec 2006. [Online]. Available:
  \url{https://doi.org/10.1016/S1672-6529(07)60003-1}
\BIBentrySTDinterwordspacing

\bibitem{Suomela2005BallShapedRA}
J.~Suomela and T.~Ylikorpi, ``Ball-shaped robots: An historical overview and
  recent developments at tkk,'' in \emph{FSR}, 2005.

\bibitem{security}
M.~Seeman, M.~Broxvall, A.~Saffiotti, and P.~Wide, ``An autonomous spherical
  robot for security tasks,'' in \emph{2006 IEEE International Conference on
  Computational Intelligence for Homeland Security and Personal Safety}, Oct
  2006, pp. 51--55.

\bibitem{6618080}
S.~Guo, M.~Li, and C.~Yue, ``Performance evaluation on land of an amphibious
  spherical mother robot in different terrains,'' in \emph{2013 IEEE
  International Conference on Mechatronics and Automation}, Aug 2013, pp.
  1173--1178.

\bibitem{rotundus}
V.~Kaznov and M.~Seeman, ``Outdoor navigation with a spherical amphibious
  robot,'' in \emph{2010 IEEE/RSJ International Conference on Intelligent
  Robots and Systems}, Oct 2010, pp. 5113--5118.

\bibitem{robotics1010003}
\BIBentryALTinterwordspacing
R.~Chase and A.~Pandya, ``A review of active mechanical driving principles of
  spherical robots,'' \emph{Robotics}, vol.~1, no.~1, pp. 3--23, 2012.
  [Online]. Available: \url{http://www.mdpi.com/2218-6581/1/1/3}
\BIBentrySTDinterwordspacing

\bibitem{nizam}
K.~Nizam, ``Dynamics control of pendulums driven spherical robot,''
  \emph{Applied Mechanics and Materials}, vol. 315, pp. 192--195, 04 2013.

\bibitem{sclater2007mechanisms}
\BIBentryALTinterwordspacing
N.~Sclater and N.~Chironis, \emph{Mechanisms and Mechanical Devices Sourcebook,
  Fourth Edition}.\hskip 1em plus 0.5em minus 0.4em\relax Mcgraw-hill, 2007.
  [Online]. Available: \url{https://books.google.com/books?id=iI1UAAAAMAAJ}
\BIBentrySTDinterwordspacing

\bibitem{chainlink}
\BIBentryALTinterwordspacing
Serapid, ``Basics of rigid-chain actuators.'' [Online]. Available:
  \url{https://www.linearmotiontips.com/basics-rigid-chain-actuators/}
\BIBentrySTDinterwordspacing

\bibitem{zipchainlink}
\BIBentryALTinterwordspacing
TSUBAKI, ``Zip chain actuator, a linear actuator.'' [Online]. Available:
  \url{https://www.hfag.ch/}
\BIBentrySTDinterwordspacing

\bibitem{spiralift}
\BIBentryALTinterwordspacing
P.~Group, ``Spiralift.'' [Online]. Available:
  \url{http://www.pacospiralift.com/}
\BIBentrySTDinterwordspacing

\bibitem{7487363}
F.~Collins and M.~Yim, ``Design of a spherical robot arm with the spiral zipper
  prismatic joint,'' in \emph{2016 IEEE International Conference on Robotics
  and Automation (ICRA)}, May 2016, pp. 2137--2143.

\bibitem{Jain2009}
\BIBentryALTinterwordspacing
A.~Jain and C.~C. Kemp, ``El-e: an assistive mobile manipulator that
  autonomously fetches objects from flat surfaces,'' \emph{Autonomous Robots},
  vol.~28, no.~1, p.~45, Sep 2009. [Online]. Available:
  \url{https://doi.org/10.1007/s10514-009-9148-5}
\BIBentrySTDinterwordspacing

\bibitem{pr2}
\BIBentryALTinterwordspacing
W.~garage, ``Pr2 robot.'' [Online]. Available:
  \url{http://www.willowgarage.com/pages/pr2/overview}
\BIBentrySTDinterwordspacing

\bibitem{6766556}
R.~{Janssen}, E.~{van Meijl}, D.~{Di Marco}, R.~{van de Molengraft}, and
  M.~{Steinbuch}, ``Integrating planning and execution for ros enabled service
  robots using hierarchical action representations,'' in \emph{2013 16th
  International Conference on Advanced Robotics (ICAR)}, Nov 2013, pp. 1--7.

\bibitem{qwerty}
\BIBentryALTinterwordspacing
L.~A. Mateos, ``Red1 humanoid robot.'' [Online]. Available:
  \url{http://www.particlerobots.com/redi/design.html}
\BIBentrySTDinterwordspacing

\bibitem{s1}
A.~{Halme}, T.~{Schonberg}, and {Yan Wang}, ``Motion control of a spherical
  mobile robot,'' in \emph{Proceedings of 4th IEEE International Workshop on
  Advanced Motion Control - AMC '96 - MIE}, vol.~1, March 1996, pp. 259--264
  vol.1.

\bibitem{s2}
{Qiang Zhan}, {Yao Cai}, and {Caixia Yan}, ``Design, analysis and experiments
  of an omni-directional spherical robot,'' in \emph{2011 IEEE International
  Conference on Robotics and Automation}, May 2011, pp. 4921--4926.

\bibitem{s3}
A.~{Bicchi}, A.~{Balluchi}, D.~{Prattichizzo}, and A.~{Gorelli}, ``Introducing
  the "sphericle": an experimental testbed for research and teaching in
  nonholonomy,'' in \emph{Proceedings of International Conference on Robotics
  and Automation}, vol.~3, April 1997, pp. 2620--2625 vol.3.

\bibitem{s4}
C.~{Camicia}, F.~{Conticelli}, and A.~{Bicchi}, ``Nonholonomic kinematics and
  dynamics of the sphericle,'' in \emph{Proceedings. 2000 IEEE/RSJ
  International Conference on Intelligent Robots and Systems (IROS 2000) (Cat.
  No.00CH37113)}, vol.~1, Oct 2000, pp. 805--810 vol.1.

\bibitem{s5}
R.~{Mukherjee}, M.~A. {Minor}, and J.~T. {Pukrushpan}, ``Simple motion planning
  strategies for spherobot: a spherical mobile robot,'' in \emph{Proceedings of
  the 38th IEEE Conference on Decision and Control (Cat. No.99CH36304)},
  vol.~3, Dec 1999, pp. 2132--2137 vol.3.

\bibitem{s6}
{Amir Homayoun Javadi A} and P.~{Mojabi}, ``Introducing august: a novel
  strategy for an omnidirectional spherical rolling robot,'' in
  \emph{Proceedings 2002 IEEE International Conference on Robotics and
  Automation (Cat. No.02CH37292)}, vol.~4, May 2002, pp. 3527--3533 vol.4.

\bibitem{s7}
S.~{Bhattacharya} and S.~K. {Agrawal}, ``Design, experiments and motion
  planning of a spherical rolling robot,'' in \emph{Proceedings 2000 ICRA.
  Millennium Conference. IEEE International Conference on Robotics and
  Automation. Symposia Proceedings (Cat. No.00CH37065)}, vol.~2, April 2000,
  pp. 1207--1212 vol.2.

\bibitem{urchin}
\BIBentryALTinterwordspacing
L.~A. Mateos, ``Bionic sea urchin with foldable telescopic actuator,
  https://arxiv.org/abs/2001.04250.'' [Online]. Available:
  \url{https://arxiv.org/abs/2001.04250}
\BIBentrySTDinterwordspacing

\bibitem{videos}
\BIBentryALTinterwordspacing
------, ``Articulated rigid rack how it works and videos,
  https://www.particlerobots.com/luismateos/particle/.'' [Online]. Available:
  \url{https://www.particlerobots.com/luismateos/particle/}
\BIBentrySTDinterwordspacing

\bibitem{8793525}
L.~A. {Mateos}, W.~{Wang}, B.~{Gheneti}, F.~{Duarte}, C.~{Ratti}, and D.~{Rus},
  ``Autonomous latching system for robotic boats,'' in \emph{2019 International
  Conference on Robotics and Automation (ICRA)}, May 2019, pp. 7933--7939.

\end{thebibliography}

\end{document}